\documentclass[runningheads]{llncs}
\usepackage{graphicx}
\usepackage{amsmath}
\usepackage{amssymb}
\usepackage{multirow}
\usepackage{array}
\usepackage[misc,geometry]{ifsym}
\begin{document}
\title{Bidirectional Regression for Arbitrary-Shaped Text Detection
}
%
%
\author{Tao Sheng \orcidID{0000-0002-1152-8316} \and
Zhouhui Lian  (\Letter)  \orcidID{0000-0002-2683-7170}}
\authorrunning{T Sheng and Z Lian}
%
\institute{Wangxuan Institute of Computer Technology, Peking University, Beijing, China
\email{\{shengtao,lianzhouhui\}@pku.edu.cn}}
\maketitle              
\begin{abstract}
	Arbitrary-shaped text detection has recently attracted increasing 
	interests and witnessed rapid development with the popularity of 
	deep learning algorithms. Nevertheless, existing approaches often 
	obtain inaccurate detection results, mainly 
	due to the relatively weak ability to utilize context 
	information and the inappropriate choice of offset references. 
	This paper presents a novel text instance expression which integrates 
	both foreground and background information into the pipeline, and 
	naturally uses the pixels near text boundaries as the offset 
	starts. Besides, a corresponding post-processing algorithm is also designed  
	to sequentially combine the four prediction results and reconstruct the 
	text instance accurately. We evaluate our method on several 
	challenging scene text benchmarks, including both curved and 
	multi-oriented text datasets. Experimental results demonstrate that the 
	proposed approach obtains superior or competitive performance 
	compared to other state-of-the-art methods, e.g., 83.4$\%$ F-score for 
	Total-Text, 82.4$\%$ F-score for MSRA-TD500, etc.

\end{abstract}


\section{Introduction}

In many text-related applications, robustly detecting scene texts with 
high accuracy, namely localizing the bounding box or region of each 
text instance with high Intersection over Union (IoU) to the ground 
truth, is fundamental and crucial to the quality of service. For example, 
in vision-based translation applications, the process of generating 
clear and coherent translations is highly dependent on the text detection 
accuracy. However, due to the variety of text scales, shapes and 
orientations, and the complex backgrounds in natural images, scene text 
detection is still a tough and challenging task.

\begin{figure}
\begin{center}
   \includegraphics[width=0.5\linewidth]{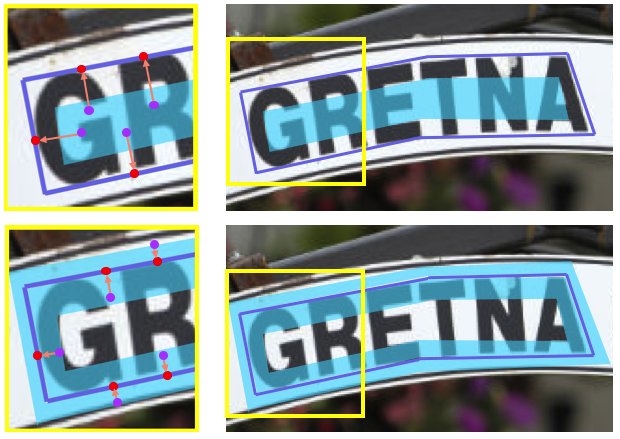}
\end{center}
   \caption{The comparison of detection results of MSR~\cite{msr} and our method. In the first row, 
   MSR predicts central text regions (top right) and distance maps 
   according to the distance between each predicted text pixel and 
   its nearest text boundary (top left). In the second row, we choose 
   the pixels around the text boundary for further regression. 
   The pixels for regression are marked as light blue.}
\label{fig:compareMSR}
\end{figure}

With the rapid development of deep convolutional neural networks (DCNN), 
a number of effective methods ~\cite{east,ctw1500,msr,spcnet,pse,textray} have been proposed for detecting texts in scene images, 
achieving promising performance. Among all these 
DCNN-based approaches, the majority of them can be roughly classified 
into two categories: regression-based methods with anchors and segmentation-based methods.
Regression-based methods are typically motivated by generic object detection~\cite{fasterrcnn,ssd,maskrcnn,cascadercnn}, and treat text 
instances as a specific kind of object. However, it is difficult to manually 
design appropriate anchors for irregular texts. 
On the contrary, segmentation-based methods prefer to regard scene text detection 
as a segmentation task~\cite{fcn,deeplab,dingetal,panet} and need extra predictions besides segmentation 
results to rebuild text instances. Specifically, as shown in Fig.~\ref{fig:compareMSR}, 
MSR~\cite{msr} predicts central text regions and distance maps according to the 
distance between each predicted text pixel and its nearest text boundary, 
which brings the state-of-the-art scene text detection accuracy. Nevertheless, 
the performance of MSR is still restricted by its relatively weak capability 
to utilize the context information around the text boundaries. That is 
to say, MSR only uses the pixels inside the central areas for detection, 
while ignores the pixels around the boundaries which include necessary 
context information. Moreover, the regression from central text pixels makes the 
positions of predicted boundary points ambiguous because there is a huge 
gap between them, requiring the network to have a large receptive field.

To solve these problems, we propose an arbitrary-shaped text detector, 
which makes full use of context information around the text boundaries 
and predicts the boundary points in a natural way. Our 
method contains two steps: 1) input images are feed to the convolutional 
network to generate text regions, text kernels, pixel offsets, and pixel 
orientations. 2) at the stage of post-processing, the previous predictions 
are fused one by one to rebuild the final results. Unlike MSR, we choose 
the pixels around the text boundaries for further regression (Fig.~\ref{fig:compareMSR} bottom). 
Due to the regression computed in two directions, either from the external 
or the internal text region to the text boundary, we call our method Bidirectional Regression. 
For a fair comparison, we choose a simple and commonly-used network 
structure as the feature extractor,
ResNet50~\cite{resnet} for backbone and FPN~\cite{fpn} for detection neck. Then, the segmentation 
branch predicts text regions and text kernels, while the regression branch 
predicts pixel offsets and pixel orientations which help to separate adjacent 
text lines. Also, we propose a novel post-processing algorithm 
to reconstruct final text instances accurately. We conduct extensive 
experiments on four challenging benchmarks including Total-Text~\cite{totaltext}, CTW1500~\cite{ctw1500}, 
ICDAR2015~\cite{icdar2015}, and MSRA-TD500~\cite{msratd500}, which demonstrate that our method obtains 
superior or comparable performance compared to the state of the art.

Major contributions of our work can be summarized as follows:
\begin{itemize}
	\item [1)] To overcome the drawbacks of MSR, we propose a new kind 
	of text instance expression which makes full use of context 
	information and implements the regression in a natural manner.
	\item [2)] To get complete text instances robustly, we develop a novel 
	post-processing algorithm that sequentially combines the predictions to get accurate text instances.
	\item [3)] The proposed method achieves superior or comparable performance
	compared to other existing approaches on two curved text benchmarks and two 
	oriented text benchmarks.
\end{itemize}


\section{Related Work}

Convolutional neural network approaches have recently been very successful 
in the area of scene text detection. CNN-based scene text detectors can be 
roughly classified into two categories: regression-based methods and
segmentation-based methods.

\textbf{Regression-based detectors} usually inherit from generic object 
detectors, such as Faster R-CNN~\cite{fasterrcnn} and SSD~\cite{ssd}, 
directly regressing the bounding 
boxes of text instances. TextBoxes~\cite{textboxes} adjusts the aspect ratios of anchors 
and the scales of convolutional kernels of SSD to deal with the significant 
variation of scene texts. TextBoxes++~\cite{textboxes++} and EAST~\cite{east} further regress quadrangles 
of multi-oriented texts in pixel level with and without anchors, respectively. 
For better detection of long texts, RRD~\cite{rrd} generates rotation-invariant features 
for classification and rotation-sensitive features for regression. RRPN~\cite{rrpn} 
modifies Faster R-CNN by adding rotation to proposals for titled text detection. 
The methods mentioned above achieve excellent results in several benchmarks. 
Nevertheless, most of them suffer from the complex anchor settings or the 
inadequate description of irregular texts.

\textbf{Segmentation-based detectors} prefer to treat scene text detection 
as a semantic segmentation problem and apply a matching post-processing 
algorithm to get the final polygons. Zhang et al.~\cite{zhangetal} utilized FCN~\cite{fcn} to estimate 
text regions and further distinguish characters with MSER~\cite{mser}. In PixelLink~\cite{pixellink}, 
text/non-text and links predictions in pixel level are carried out to
separate adjacent text instances. Chen et al.~\cite{chenetal} proposed the concept of
attention-guided text border for better training. SPCNet~\cite{spcnet} and Mask TextSpotter~\cite{masktextspotter}
adopt the architecture of Mask R-CNN~\cite{maskrcnn} in the instance segmentation task 
to detect the texts with arbitrary shapes. In TextSnake~\cite{textsnake}, text instances 
are represented with text center lines and ordered disks. 
MSR~\cite{msr} predicts central text regions and distance 
maps according to the distance between each predicted text pixel and 
its nearest text boundary. PSENet~\cite{pse} proposes progressive scale expansion 
algorithm, learning text kernels with multiple scales. However, these 
methods all lack the rational utilization of context information, and 
thus often result in inaccurate text detection.

Some previous methods~\cite{chenetal,masktextspotter,textfield} also try to 
strengthen the utilization of context information according to the inference 
of the text border map or the perception of the whole text polygon. Regrettably, 
they only focus on pixels inside the text polygon and miss pixels outside the label.
Different from existing methods, a unique text expression is proposed 
in this paper to force the network to extract foreground and background 
information simultaneously and produce a more representative feature 
for precise localization.


\section{Methodology}

In this section, we first compare the text instance expressions of 
common object detection models, a curved text detector MSR~\cite{msr} and ours. Then, we describe the
whole network architecture of our proposed method. Afterwards, we 
elaborate on the post-processing algorithm and the generation procedure of
arbitrary-shaped text instances. Finally, the details of the loss 
function in the training phase are given.

\subsection{Text Instance Expression}

\begin{figure*}[t]
\begin{center}
	\includegraphics[width=1.0\linewidth]{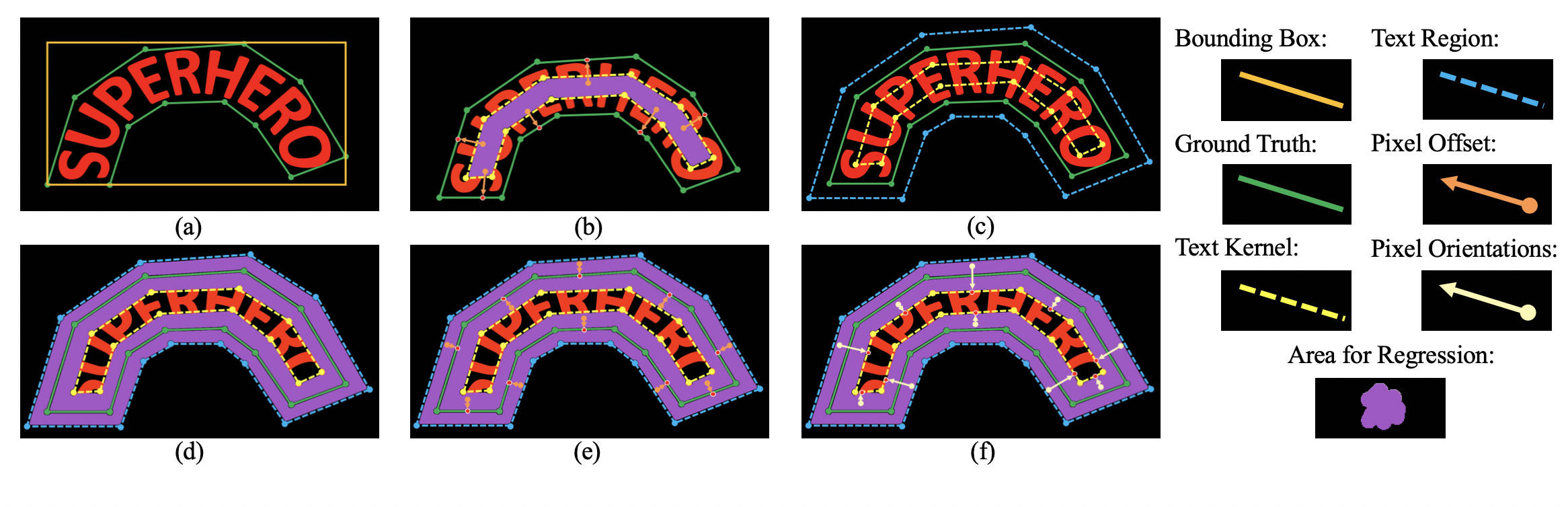}
\end{center}
   \caption{The comparison of three text instance expressions:
   (a) bounding boxes, (b) MSR's expression~\cite{msr}, 
   and (c-f) our Bidirectional Regression.}
\label{fig:expression}
\end{figure*}

\textbf{Bounding Box} \text{ } A robust scene text detector must 
have a well-defined expression for text instances. In generic object 
detection methods, the text instances 
are always represented as bounding boxes, namely 
rotated rectangles or quadrangles, whose shapes 
heavily rely on vertices or geometric centers. However, it is difficult 
to decide vertices or geometric center of a curved text instance, 
especially in natural images. Besides, as shown in Fig.~\ref{fig:expression}(a), 
the bounding box (orange solid line) can not fit the boundary 
(green solid line) of the curved text well and introduces a large 
number of background noises, which could be problematic for detecting 
scene texts. Moreover, at the stage of scene text 
recognition, the features extracted in this way may confuse the 
model to obtain incorrect recognition results.

\textbf{MSR} \text{ } does not have this expression problem because as 
shown in Fig.~\ref{fig:expression}(b), it predicts central text 
regions (violet area) and offset maps according to the distance 
(orange line with arrow) between each predicted text pixel and 
its nearest text boundary. Note that the central text regions 
are derived from the original text boundary (green solid line), 
which help to separate adjacent words or text lines. Obviously, 
a central text region $S_{kernel}$ can be discretized into a 
set of points $\{p_1,p_2,\dots,p_T\}$, and $T$ is the 
number of points. Naturally, the offset between $p_i$ 
and its nearest boundary can be represented as 
$\{\Delta p_i|i=1,2,\dots,T\}$. Further, the contour of the 
text instance can be represented with the point set,
\begin{equation}
R=\{p_i+\Delta p_i|i=1,2,\dots,T\}.
\label{MSR}
\end{equation}
We observe that MSR focuses on the location of center points 
and the relationship between center and boundary points, 
while ignores the crucial context information around the 
text boundaries and chooses inappropriate references as the 
starts of offsets. More specifically, 1) the 
ignored features around the boundary are the most discriminative 
part of the text instance's features extracted from the whole image, because foreground
information inside the boundary and background information 
outside the boundary are almost completely different and 
easy to be distinguished. 2) compared to the points around the 
boundary, the center ones are farther away from the boundary, 
which can not give enough evidences to decide the position of
the scene text instance. In other words, if we can utilize
the context information well and choose better references,
we will possibly be able to address MSR's limitations and 
further improve the detection performance.

\textbf{Bidirectional Regression (Ours)} \text{ } We now define our
text instance expression so that the generated feature maps can have more useful
contextual information for detection. We use four predictions
to represent a text instance, including the text region, text 
kernel, pixel offset, and pixel orientation. As shown in 
Fig.~\ref{fig:expression}(c), inspired by PSENet~\cite{pse}, the text 
kernel is generated by shrinking the annotated polygon $P$
(green solid line) to the yellow dotted line using the Vatti 
clipping algorithm~\cite{vatti}. The offset $d$ of shrinking is computed 
based on the perimeter and area of the original polygon $P$:
\begin{equation}
	d=\dfrac{\textrm{Area}(P)\times(1-\alpha^2)}{\textrm{Perimeter}(P)},
\end{equation}
where $\alpha$ is the shrink ratio, set to 0.6 empirically.
To contain the background information, we enlarge the  
annotated polygon $P$ to the blue dotted line in the same way 
and produce the text region. The expansion ratio 
$\beta$ is set to 1.2 in this work. Similarly, the text 
kernel $S_{kernel}$ and the text region $S_{text}$ can be 
treated as point sets, and further the text border 
(violet area in Fig.~\ref{fig:expression}(d)) is the 
difference set of both, which can be formulated as:
\begin{equation}
	S_{border}=S_{text}-S_{kernel},
\end{equation}
For convenience, we use $\{p_i|i=1,2,\dots,T'\}$ to represent
the text border, where $T'$ is the number of points in 
this area. As shown in Fig.~\ref{fig:expression}(e) and Fig.~\ref{fig:expression}(f), 
we establish the pixel offset map according to the distance 
(orange line with arrow) between each text border point and its
nearest text boundary, and the pixel orientation map according 
to the orientation (white line with arrow) from each text 
border point to its nearest text kernel point. Note that 
if two instances overlap, the smaller one has higher priority. 
Like MSR, we 
use sets $\{\Delta p_i|i=1,2,\dots,T'\}$ and $\{\overrightarrow{\theta_i}|i=1,2,\dots,T'\}$
to represent the pixel offset and the pixel orientation 
respectively, where $\overrightarrow{\theta_i}$ is a unit 
vector. Similar to Eq.~\ref{MSR}, the final predicted contour 
can be formulated as follows:
\begin{equation}
\begin{aligned}
R=\{p_i+\Delta p_i|\exists p_k\in S_{kernel}, D(p_i,p_k)<\gamma \textrm{ and } \\\dfrac{\overrightarrow{p_i p_k}}{|p_ip_k|} \cdot \overrightarrow {\theta_i} > \epsilon, i=1,2,\dots,T'\},
\end{aligned}
\end{equation}
where $D(\cdot)$ means the Euclidean distance between two 
points. $\gamma$ and $\epsilon$ are constants and we choose
$\gamma=3.0, \epsilon=\cos(25^{\circ})\approx 0.9063$ empirically. That is to say, 
if there exists a text border point that is close enough to the 
text kernel point, and the vector composed of these two 
points has a similar direction with the predicted pixel orientation,
then the text border point will be shifted according to the 
amount of the predicted pixel offset, and be treated 
as the boundary point of the final text instance.

Comparing the areas for regression in MSR and our approach
(Fig.~\ref{fig:expression}(b)(e)), we can see that our proposed 
method chooses the points much closer to the boundary so that 
we do not require the network to have large receptive 
fields when detecting large text instances. Meanwhile, 
our model learns the foreground and background features
macroscopically, and mixes them into a more discriminating
one to localize the exact position of scene texts. To sum up, 
we design a powerful expression for 
arbitrary-shaped text detection. 

\subsection{Network Architecture}

\begin{figure*}[t]
\begin{center}
	\includegraphics[width=1.0\linewidth]{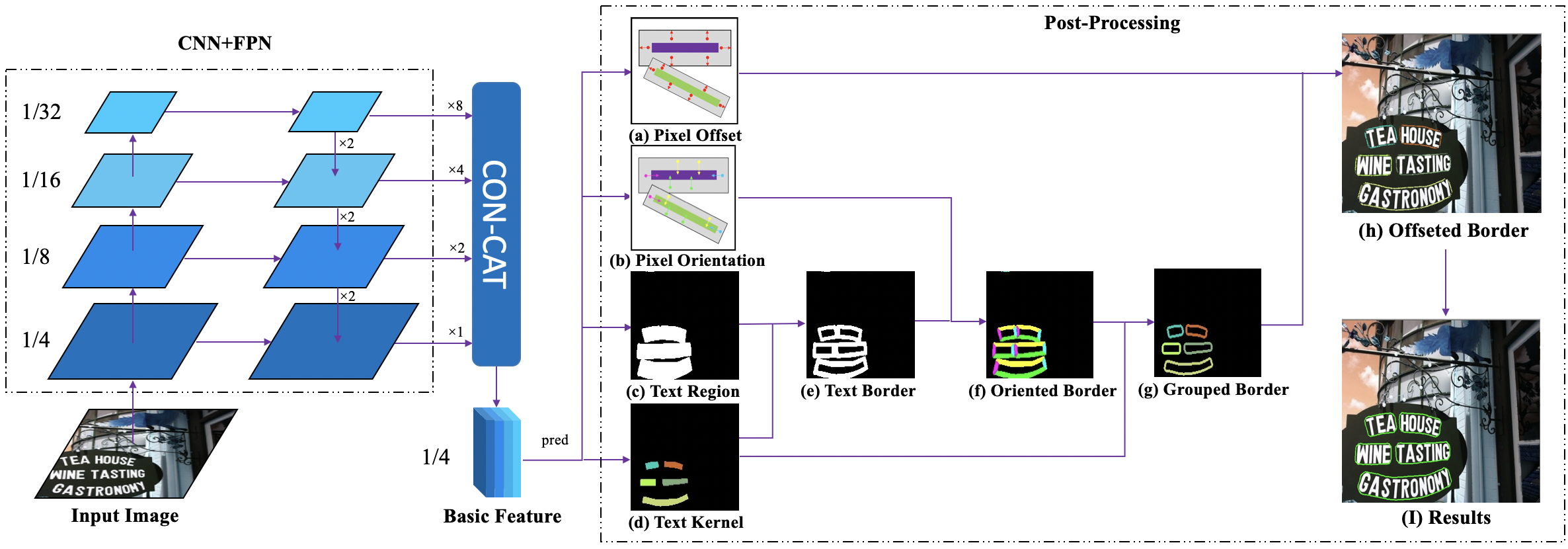}
\end{center}
   \caption{An overview of our proposed model. Our method
   contains two components, the CNN-based network and 
   the post-processing algorithm. (a) We use ResNet-50
   and FPN to extract the feature pyramid, and concatenate
   them into a basic feature for further prediction. 
   (b) The post-processing algorithm takes two forecasts 
   as inputs and produces a new one every 
   step, reconstructing the scene texts with arbitrary 
   shapes finally.}
\label{fig:pipeline}
\end{figure*}

From the network's perspective, our model is surprisingly 
simple, and Fig.~\ref{fig:pipeline} illustrates the whole 
architecture. For a fair comparison with MSR, we employ 
ResNet-50~\cite{resnet} as the backbone network to extract initial 
multi-scale features from input images. A total of 4 feature 
maps are generated from the Res2, Res3, Res4, and Res5 layers of the 
backbone, and they have strides of 4, 8, 16, 32 pixels
with respect to the input image respectively. To reduce
the computational cost and the network complexity, we use 
$1\times1$ convolutions to reduce the channel number of 
each feature map to 256. Inspired by FPN~\cite{fpn}, the feature 
pyramid is enhanced by gradually merging the features of
adjacent scales in a top-down manner. Then, the deep but
thin feature maps are fused by bilinear interpolation 
and concatenation into a basic feature, whose stride is
4 pixels and the channel number is 1024. The basic feature
is used to predict text regions, text kernels, pixel offsets, 
and pixel orientations simultaneously. Finally, we apply a
sequential and efficient post-processing algorithm to obtain
the final text instances.

\subsection{Post-Processing}

As described in the above subsection and illustrated in 
Fig.~\ref{fig:pipeline}(a-d), four predictions are 
generated from the basic feature, followed by the 
post-processing algorithm. The text region can describe the text 
instance coarsely but can not separate two adjacent instances 
(Fig.~\ref{fig:pipeline}(c)). In contrast, the text kernel 
can separate them but can not describe them 
(see Fig.~\ref{fig:pipeline}(d)). Therefore, we use text kernels
to determine the coarse position, then use pixel orientations 
to classify the ungrouped text region points, and use pixel 
offsets to slightly modify the contours.

We first find the connected components in the text kernel
map, and each connected component represents the kernel 
of a single text instance. For better visualization, different 
kernels are painted with different colors (see Fig.~\ref{fig:pipeline}(d)). 
Moreover, the pixel offset map and the pixel orientation 
map are difficult to be visualized, so we replace the real 
ones with the diagrammatic sketches in a fictitious scene (see Fig.~\ref{fig:pipeline}(a)(b)).
Then, we combine the text region and the text kernel, 
obtaining the text border (see Fig.~\ref{fig:pipeline}(e)), 
which is the difference set of two predictions and meanwhile the 
aggregation of the ungrouped text region points. Combined 
with the pixel orientation, each text border point has
its own orientation. As shown in Fig.~\ref{fig:pipeline}(f), four 
colors (yellow, green, violet, and blue) represent the four directions (up, down, left, and right)
correspondingly. Afterwards, the oriented border and 
the text kernel are combined together to classify the 
text border points into the groups of previously connected 
components in the text kernel (see Fig.~\ref{fig:pipeline}(g)) 
according to the difference between the predicted orientation 
and the orientation from each text border point to its nearest text 
kernel point. A text border point should be 
deserted if the distance to its nearest text kernel point 
is too far. Furthermore, each grouped border point will be 
shifted to its nearest point on the text boundary, which can 
be calculated by summing up the coordinates of the point and 
the predicted offset in the pixel offset map 
(see Fig.~\ref{fig:pipeline}(h)). Finally, we adopt the 
Alpha-Shape Algorithm to produce concave polygons enclosing 
the shifted points of each text kernel group 
(see Fig.~\ref{fig:pipeline}(I)), precisely reconstructing 
the shapes of the text instances in scene images. Through 
this effective post-processing algorithm, we can detect 
scene texts with arbitrary shapes fast and accurately, 
which is experimentally proved in Section~\ref{sec:experiments}.

\subsection{Loss Function}

Our loss function can be formulated as:
\begin{equation}
\mathcal{L}=\mathcal{L}_{text}+\lambda_1\mathcal{L}_{kernel}+\lambda_2(\mathcal{L}_{offset}+\mathcal{L}_{orientation}),
\end{equation}
where $\mathcal{L}_{text}$ and $\mathcal{L}_{kernel}$ 
denote the binary segmentation loss of text regions 
and text kernels respectively, $\mathcal{L}_{offset}$ 
denotes the regression loss of pixel offsets, 
and $\mathcal{L}_{orientation}$ denotes the orientation loss 
of pixel orientations. $\lambda_1$ and $\lambda_2$ 
are normalization constants to balance the weights 
of the segmentation and regression loss. 
We set them to 0.5 and 0.1 in all experiments.

The prediction of text regions and text kernels is basically 
a pixel-wise binary classification problem. We 
follow MSR and adopt the dice loss~\cite{dice} for this part. Considering 
the imbalance of text and non-text pixels in the text regions,
Online Hard Example Mining (OHEM)~\cite{ohem} is also adopted to select the 
hard non-text pixels when calculating $\mathcal{L}_{text}$.

The prediction of the distance from each
text border point to its nearest text boundary is a 
regression problem. Following the regression for bounding 
boxes in generic object detection, we use the Smooth L1 loss~\cite{fastrcnn} 
for supervision, which is defined as:
\begin{equation}
\mathcal{L}_{offset}=\dfrac{1}{|S_{border}|} \sum_i \textrm{Smooth}_\textrm{L1}(\Delta p_i-\Delta p_i^*),
\end{equation}
where $\Delta p_i$ and $\Delta p_i^*$ denote the predicted 
offset and the corresponding ground truth, respectively, $\textrm{Smooth}_\textrm{L1}(\cdot)$ 
denotes the standard Smooth L1 loss, and $|S_{border}|$ 
denotes the number of text border points. Moreover, 
the prediction of the orientation from each text border 
point to its nearest text kernel point can also be treated
as a regression problem. For simplicity, we adopt the cosine 
loss defined as follows:
\begin{equation}
\mathcal{L}_{orientation}=\dfrac{1}{|S_{border}|} \sum_i\left(1-\overrightarrow{\theta_i}\cdot\overrightarrow{\theta_i^*}\right),
\end{equation}
where $\overrightarrow{\theta_i}\cdot\overrightarrow{\theta_i^*}$
denote the dot product of the predicted direction vector and
its ground truth, which is equal to the cosine value of the angle between these two orientations. 
Note that we only take the points in the text border into 
consideration when calculating $\mathcal{L}_{offset}$ and 
$\mathcal{L}_{orientation}$.


\section{Experiments}
\label{sec:experiments}

\subsection{Datasets}

\textbf{SynthText}~\cite{synth} is a synthetical dataset containing 
more than 800,000 synthetic scene text images, most of 
which are annotated at word level with multi-oriented 
rectangles. We pre-train our model on this dataset. \\
\textbf{Total-Text}~\cite{totaltext} is a curved text dataset which contains 
1,255 training images and 300 testing images. The texts are 
all in English and contain a large number of horizontal, 
multi-oriented, and curved text instances, each of which 
is annotated at word level with a polygon. \\
\textbf{CTW1500}~\cite{ctw1500} is another curved text dataset that 
has 1,000 images for training and 500 images for testing.
The dataset focuses on curved texts, which are largely 
in English and Chinese, and annotated at text-line level 
with 14-polygons. \\
\textbf{ICDAR2015}~\cite{icdar2015} is a commonly-used dataset for scene text 
detection, which contains 1,000 training images and 
500 testing images. The dataset is captured by Google 
Glasses, where text instances are annotated at word 
level with quadrilaterals. \\
\textbf{MSRA-TD500}~\cite{msratd500} is a small dataset that contains 
a total of 500 images, 300 for training and the remaining 
for testing. All captured text instances are in English 
and Chinese, which are annotated at text-line level with 
best-aligned rectangles. Due to the rather small scale of 
the dataset, we follow the previous works~\cite{east,textsnake} to add the 
400 training images from HUST-TR400~\cite{husttr400} into the training set.

\subsection{Implementation Details}

The following settings are used throughout the experiments.
The proposed method is implemented with the deep learning
framework, Pytorch, on a regular GPU workstation with 4
Nvidia Geforce GTX 1080 Ti. For the network architecture, 
we use the ResNet-50~\cite{resnet} pre-trained on ImageNet~\cite{imagenet} as our backbone. 
For learning, we train our model with the batch size of 16 
on 4 GPUs for 36K iterations. Adam optimizer with a starting 
learning rate of $10^{-3}$ is used for optimization. We use the ``poly" 
learning rate strategy~\cite{zhaoetal}, where the initial rate is multiplied by 
$(1-\frac{\textrm{iter}}{\textrm{max\_iter}})^{\textrm{power}}$, 
and the ``power" is set to 0.9 in all experiments. For data 
augmentation, we apply random scale, random horizontal flip, 
random rotation, and random crop on training images. We ignore
the blurred texts labeled as DO NOT CARE in all datasets.
For others, Online hard example mining (OHEM) is used to 
balance the positive and negative samples, and the 
negative-positive ratio is set to 3.
We first pre-train our model on SynthText, and 
then fine-tune it on other datasets. The training settings 
of the two stages are the same.

\subsection{Ablation Study}

To prove the effectiveness of our proposed expression for text instances, 
we carry out ablation 
studies on the curved text dataset Total-Text. Note that, 
all the models in this subsection are pre-trained on 
SynthText first. The quantitative results of the same 
network architecture with different expressions (with corresponding 
post-processing algorithms) are shown in Tab.~\ref{tab:ab1}.

\begin{table}
\begin{center}
	\begin{tabular}{|c|c|c|c|c|}
	\hline
	\textbf{Expression} & $\beta$ & \textbf{Precision} & \textbf{Recall} & \textbf{F-score} \\
	\hline
	\hline
	\textbf{MSR} & - & 84.7 & 77.3 & 80.8 \\
	\hline
	\textbf{Ours} & 1.0 & 85.8 & 79.1 & 82.3 \\
	\hline
	\textbf{Ours} & 1.2 & \textbf{87.0} & \textbf{80.1} & \textbf{83.4} \\
	\hline
	\end{tabular}
\end{center}
	\caption{The results of models with different expressions 
	over the curved text dataset Total-Text. ``$\beta$" means 
	the expansion ratio of the text region.}
\label{tab:ab1}
\end{table}

\begin{table}
\begin{center}
	\begin{tabular}{|p{2.1cm}<{\centering}|p{1cm}<{\centering}|p{1cm}<{\centering}|p{1cm}<{\centering}|p{1cm}<{\centering}|p{1cm}<{\centering}|p{1cm}<{\centering}|}
	\hline
	\textbf{Expression} & \textbf{F} & $F_{0.5}$ & $F_{0.6}$ & $F_{0.7}$ & $F_{0.8}$ & $F_{0.9}$ \\
	\hline
	\hline
	\textbf{MSR} & 80.8 & \textbf{88.2} & 85.5 & 76.4 & 48.2 & 5.7 \\
	\hline
	\textbf{Ours} & \textbf{83.4} & 87.9 & \textbf{85.6} & \textbf{78.4} & \textbf{56.9} & \textbf{15.2} \\
	\hline
	\end{tabular}
\end{center}
	\caption{The results of models with different expressions 
	over the varying IoU thresholds from 0.5 to 0.9. ``$F_x$" 
	means the IoU threshold is set to ``$x$" when evaluating. 
	The measure of ``F" follows the Total-Text dataset, 
	setting TR to 0.7 and TP to 0.6 threshold for a fairer 
	evaluation.}
\label{tab:ab2}
\end{table}

To better analyze the capability of the proposed expression, 
we replace our text instance expression in the proposed text detector with MSR's. The 
F-score of the model with the MSR's expression (the 
first row in Tab.~\ref{tab:ab1}) drops 2.6$\%$ compared to 
our method (the third row in Tab.~\ref{tab:ab1}), which 
indicates the effectiveness of our text instance expression 
clearly. To prove the necessity of introducing 
background pixels around the text boundary,
we adjust the expansion ratio $\beta$ from 1.2 to 1.0 (the 
second row in Tab.~\ref{tab:ab1}). We can see that the 
F-score value increases by 1.1$\%$ when the model extracts the 
foreground and background features macroscopically. 
Furthermore, to judge whether the text border pixels 
are the better references for the pixel offsets or not, 
we compare the model with MSR's expression 
and ours without enlarging text instances, and notice 
that ours makes about 1.5$\%$ improvement on F-score. We 
further analyze the detection accuracy between the model 
with MSR's expression and ours by varying the evaluation 
IoU threshold from 0.5 to 0.9. Tab.~\ref{tab:ab2} shows 
that our method defeats the competitor for most IoU 
settings, especially in high IoU levels, indicating that 
our predicted polygons fit text instances better.

\subsection{Comparisons with State-of-the-Art Methods}

\begin{figure*}[t]
\begin{center}
   \includegraphics[width=1.0\linewidth]{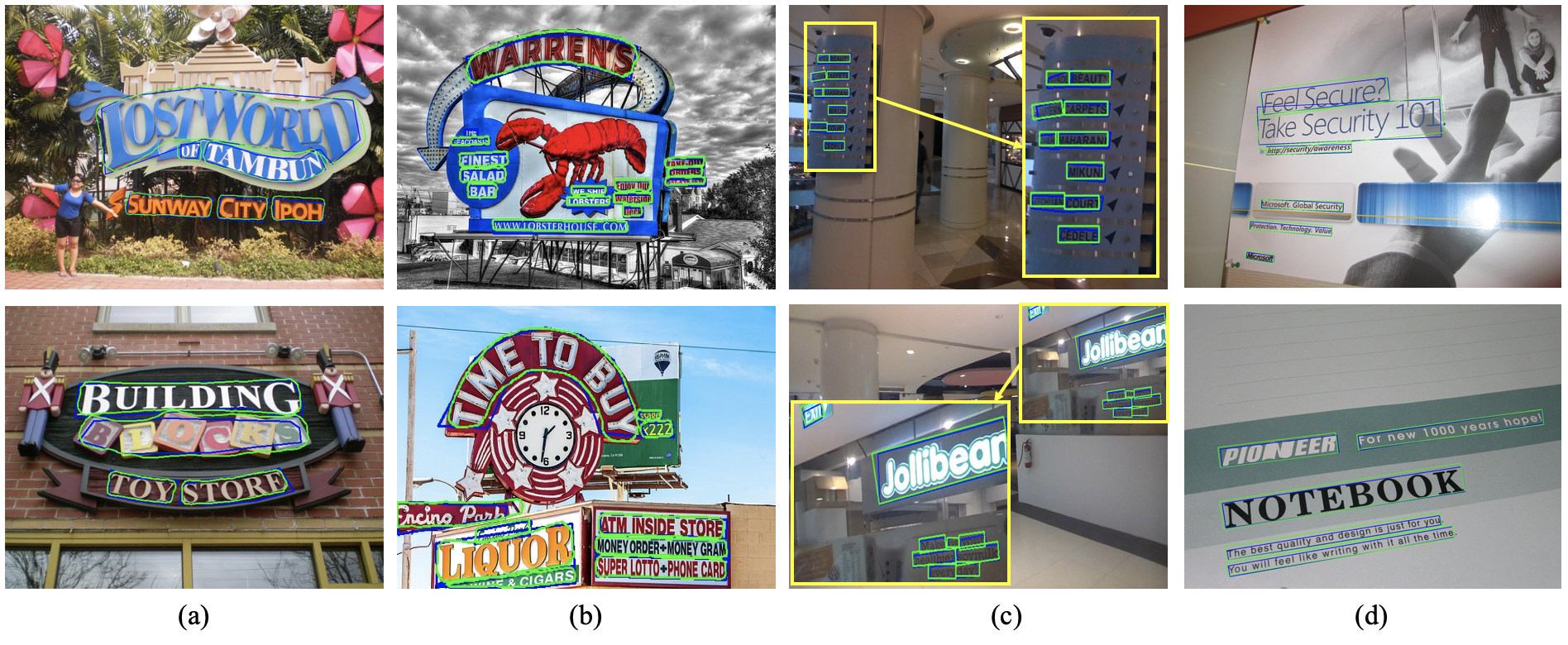}
\end{center}
   \caption{The qualitative results of the proposed method. 
   Images in columns (a)-(d) are sampled from the 
   datasets Total-Text, CTW1500, ICDAR2015, and 
   MSRA-TD500 respectively. The green polygons are 
   the detection results predicted by our method, 
   while the blue ones are ground-truth annotations.}
\label{fig:4dataset}
\end{figure*}

\textbf{Curved text detection} \text{ } We first evaluate our 
method over the datasets Total-Text and CTW1500 which contain 
many curved text instances. In the testing phase, we set the 
short side of images to 640 and keep their original aspect ratio. 
We show the experimental results in Tab.~\ref{tab:TTCTW}.
On Total-Text, our method achieves the F-score of 83.4\%, which 
surpasses all other state-of-the-art methods by at least 0.5\%. 
Especially, we outperform our counterpart, MSR, in F-score by 
over 4\%. Analogous results can be found on CTW1500. Our method 
obtains 81.8\% in F-score, the second-best one of all 
methods, which is only lower than PSENet~\cite{pse} but surpasses MSR by 
0.3\%. To sum up, our experiments conducted 
on these two datasets demonstrate the advantages of our method 
when detecting text instances with arbitrary shapes in complex 
natural scenes. We visualize our detection results in 
Fig.~\ref{fig:4dataset}(a)(b) for further inspection.

\begin{table*}
\begin{center}
	\begin{tabular}{|p{3.2cm}<{\centering}|p{1.3cm}<{\centering}|p{1.3cm}<{\centering}|p{1.3cm}<{\centering}|p{1.3cm}<{\centering}|p{1.3cm}<{\centering}|p{1.3cm}<{\centering}|}
		\hline
		\multirow{2}{*}{\textbf{Method}} & \multicolumn{3}{c|}{\textbf{Total-Text}} & \multicolumn{3}{c|}{\textbf{CTW1500}} \\
		\cline{2-7}
		 & \textbf{P} & \textbf{R} & \textbf{F} & \textbf{P} & \textbf{R} & \textbf{F} \\
		\hline
		\hline
		SegLink~\cite{seglink} & 30.3 & 23.8 & 26.7 & 42.3 & 40.0 & 40.8 \\
		\hline
		EAST~\cite{east} & 50.0 & 36.2 & 42.0 & 78.7 & 49.1 & 60.4 \\
		\hline
		Mask TextSpotter~\cite{masktextspotter} & 69.0 & 55.0 & 61.3 & - & - & - \\
		\hline
		TextSnake~\cite{textsnake} & 82.7 & 74.5 & 78.4 & 67.9 & \textbf{85.3} & 75.6 \\
		\hline
		CSE~\cite{cse} & 81.4 & 79.1 & 80.2 & 81.1 & 76.0 & 78.4 \\
		\hline
		TextField~\cite{textfield} & 81.2 & 79.9 & 80.6 & 83.0 & 79.8 & 81.4 \\
		\hline
		PSENet-1s~\cite{pse} & 84.0 & 78.0 & 80.9 & 84.8 & 79.7 & \textbf{82.2} \\
		\hline
		SPCNet~\cite{spcnet} & 83.0 & \textbf{82.8} & 82.9 & - & - & - \\
		\hline
		TextRay~\cite{textray} & 83.5 & 77.9 & 80.6 & 82.8 & 80.4 & 81.6 \\
		\hline
		\hline
		\textbf{MSR(Baseline)}~\cite{msr} & 83.8 & 74.8 & 79.0 & 85.0 & 78.3 & 81.5 \\
		\hline
		\textbf{Ours} & \textbf{87.0} & 80.1 & \textbf{83.4} & \textbf{85.7} & 78.2 & 81.8 \\
		\hline
	\end{tabular}
\end{center}
	\caption{Experimental results on the curved-text-line datasets Total-Text and CTW1500.}
\label{tab:TTCTW}
\end{table*}

\textbf{Oriented text detection} \text{ } Then we evaluate the proposed 
method over the multi-oriented text dataset ICDAR2015. In the 
testing phase, we set the short side of images to 736 for better 
detection. To fit its evaluation protocol, we use a minimum area 
rectangle to replace each output polygon. The performance on 
ICDAR2015 is shown in Tab.~\ref{tab:ICTD}. Our 
method achieves the F-score of 82.2\%, which is on par with MSR 
while the FPS of ours is 3 times of MSR. Indeed, MSR adopts the 
multi-scale multi-stage detection network, a particularly time-consuming 
architecture, so it is no surprise that its speed is lower than Ours.
Compared with state-of-the-art methods, our method is not as well as some 
competitors (e.g. PSENet~\cite{pse}, CRAFT~\cite{craft}, DB~\cite{DB}), but our method has 
the fastest inference speed (13.2 fps) and keeps a good balance 
between accuracy and latency. The qualitative illustrations in 
Fig.~\ref{fig:4dataset}(c) show that the proposed method can detect 
multi-oriented texts well. 

\textbf{Long straight text detection} \text{ } We also evaluate 
the robustness of our proposed method on the long straight text dataset 
MSRA-TD500. During inference, the short side of images is set 
to 736 for a fair comparison. As shown in Tab.~\ref{tab:ICTD}, our method 
achieves 82.4\% in F-score, which is 0.7\% better than MSR and comparable 
to the best-performing detectors DB and CRAFT. Therefore, our method is robust 
for detecting texts with extreme aspect ratios in complex scenarios 
(see Fig.~\ref{fig:4dataset}(d)).

\begin{table*}[t]
\begin{center}
	\begin{tabular}{|p{3.2cm}<{\centering}|p{1.2cm}<{\centering}|p{1.2cm}<{\centering}|p{1.2cm}<{\centering}|p{0.8cm}<{\centering}|p{1.2cm}<{\centering}|p{1.2cm}<{\centering}|p{1.2cm}<{\centering}|}
		\hline
		\multirow{2}{*}{\textbf{Method}} & \multicolumn{4}{c|}{\textbf{ICDAR2015}} & \multicolumn{3}{c|}{\textbf{MSRA-TD500}} \\
		\cline{2-8}
		 & \textbf{P} & \textbf{R} & \textbf{F} & \textbf{FPS} & \textbf{P} & \textbf{R} & \textbf{F} \\
		\hline
		\hline
		SegLink~\cite{seglink} & 73.1 & 76.8 & 75.0 & - & 86.6 & 70.0 & 77.0 \\
		\hline
		RRPN~\cite{rrpn} & 82.0 & 73.0 & 77.0 & - & 82.0 & 68.0 & 74.0 \\
		\hline
		EAST~\cite{east} & 83.6 & 73.5 & 78.2 & \textbf{13.2} & 87.3 & 67.4 & 76.1 \\
		\hline
		Lyu et al.~\cite{lyu2018multi} & \textbf{94.1} & 70.7 & 80.7 & 3.6 & 87.6 & 76.2 & 81.5 \\
		\hline
		DeepReg~\cite{deepreg} & 82.0 & 80.0 & 81.0 & - & 77.0 & 70.0 & 74.0 \\
		\hline
		RRD~\cite{rrd} & 85.6 & 79.0 & 82.2 & 6.5 & 87.0 & 73.0 & 79.0 \\
		\hline
		PixelLink~\cite{pixellink} & 82.9 & 81.7 & 82.3 & 7.3 & 83.0 & 73.2 & 77.8 \\
		\hline
		TextSnake~\cite{textsnake} & 84.9 & 80.4 & 82.6 & 1.1 & 83.2 & 73.9 & 78.3 \\
		\hline
		Mask TextSpotter~\cite{masktextspotter} & 85.8 & 81.2 & 83.4 & 4.8 & - & - & - \\
		\hline
		PSENet-1s~\cite{pse} & 86.9 & \textbf{84.5} & 85.7 & 1.6 & - & - & - \\
		\hline
		CRAFT~\cite{craft} & 89.8 & 84.3 & 86.9 & 8.6 & 88.2 & 78.2 & \textbf{82.9} \\
		\hline
		DB~\cite{DB} & 91.8 & 83.2 & \textbf{87.3} & 12.0 & \textbf{90.4} & 76.3 & 82.8 \\
		\hline
		\hline
		\textbf{MSR(Baseline)}~\cite{msr} & 86.6 & 78.4 & 82.3 & 4.3 & 87.4 & 76.7 & 81.7 \\
		\hline
		\textbf{Ours} & 82.6 & 81.9 & 82.2 & \textbf{13.2} & 83.7 & \textbf{81.1} & 82.4 \\
		\hline
	\end{tabular}
\end{center}
	\caption{Experimental results on the oriented-text-line dataset ICDAR2015 and long-straight-text-line dataset MSRA-TD500.}
\label{tab:ICTD}
\end{table*}


\section{Conclusion}

In this paper, we analyzed the limitations of existing segmentation-based scene text detectors and proposed a novel 
text instance expression to address these limitations. Moreover, considering their limited ability 
to utilize context information, our method extracts both foreground 
and background features for robust detection. The pixels around 
text boundaries are chosen as references of the predicted offsets 
for accurate localization. Besides, a corresponding post-processing 
algorithm is introduced to generate the final text instances. 
Extensive experiments demonstrated that our method achieves the performance superior 
or comparable to other state-of-the-art approaches on several publicly 
available benchmarks.

\subsubsection{Acknowledgements}
This work was supported by Beijing Nova Program of Science and Technology (Grant No.: Z191100001119077), Center For Chinese Font Design and Research, and Key Laboratory of Science, Technology and Standard in Press Industry (Key Laboratory of Intelligent Press Media Technology).

%
%
%
\bibliographystyle{splncs04}
\bibliography{mybibliography}
%

%
%
%
%
\end{document}